\documentclass[11pt]{article}
\usepackage{coling2018}
\usepackage{times}
\usepackage{url}
\usepackage{latexsym}
\usepackage{wrapfig}
\usepackage{array}
\usepackage{lscape}
\usepackage{longtable}
\usepackage{amsmath}
\usepackage{fancyvrb}
\usepackage{paralist}

\usepackage[usenames,dvipsnames]{xcolor}
\usepackage{tcolorbox}
\usepackage{tabularx}
\usepackage{array}
\usepackage{colortbl}
\tcbuselibrary{skins}

\newcolumntype{Y}{>{\raggedleft\arraybackslash}X}
\newcolumntype{Z}{>{\raggedright\arraybackslash}X}

\tcbset{tab1/.style={fonttitle=\large,fontupper=\normalsize,
colback=gray!10!white,colframe=red!75!black,colbacktitle=Salmon!40!white,
coltitle=black,center title,freelance,frame code={
\foreach \n in {north east,north west,south east,south west}
{\path [fill=red!75!black] (interior.\n) circle (3mm); };},}}

\tcbset{tab2/.style={enhanced,fontupper=\normalsize,
colback=gray!5!white,colframe=red!50!black,colbacktitle=Salmon!40!white,
coltitle=black,center title}}

\usepackage{tikz-qtree-compat}
\usetikzlibrary{shapes, fit, arrows.meta}

\usepackage[color=red!5]{todonotes}
\let\todonote\todo

\renewcommand{\todo}[1]{\todonote[size=\tiny]{#1}{\textcolor{red}{(TODO: #1)}}}
\newcolumntype{v}[1]{%
  >{\raggedright\hspace{0pt}}p{#1}%
}
\newcommand{\inlinetext}[1]{``\textit{#1}''}
\newcommand{\tuple}[3]{$\langle$#1; #2; #3$\rangle$}
\newcommand{\texttuple}[3]{$\langle$\textit{#1}; \textit{#2}; \textit{#3}$\rangle$}

\title{A Survey on Open Information Extraction}

\author{Christina Niklaus\textsuperscript{1}, Matthias Cetto\textsuperscript{1}, Andr\'{e} Freitas\textsuperscript{2} \and Siegfried Handschuh\textsuperscript{1} \\
  \textsuperscript{1} Faculty of Computer Science and Mathematics, University of Passau\\
  {\tt \{christina.niklaus, matthias.cetto, siegfried.handschuh\}{\tt @uni-passau.de}}\\
  \textsuperscript{2} School of Computer Science, University of Manchester\\
  {\tt andre.freitas@manchester.ac.uk}
\\}

\date{}

\begin{document}
\maketitle
\begin{abstract}
We provide a detailed overview of the various approaches that were proposed to date to solve the task of Open Information Extraction. We present the major challenges that such systems face, show the evolution of the suggested approaches over time and depict the specific issues they address. In addition, we provide a critique of the commonly applied evaluation procedures for assessing the performance of Open IE systems and highlight some directions for future work.
\end{abstract}

\section{Introduction}

%
%
\blfootnote{
    %
    %
    %
    
    \hspace{-0.65cm}  
    This work is licenced under a Creative Commons 
    Attribution 4.0 International Licence.
    Licence details:
    \url{http://creativecommons.org/licenses/by/4.0/}
     
    %
}

\label{intro}

Information extraction (IE) turns the unstructured information expressed in natural language text into a structured representation \cite{Jurafsky:2009:SLP:1214993} in the form of relational tuples consisting of a set of arguments and a phrase denoting a semantic relation between them: \texttuple{arg1}{rel}{arg2}. Traditional approaches to IE focus on answering narrow, well-defined requests over a predefined set of target relations on small, homogeneous corpora. To do so, they take as input the target relation along with hand-crafted extraction patterns or patterns learned from hand-labeled training examples (e.g., \newcite{Agichtein:2000:SER:336597.336644}, \newcite{Brin:1998:EPR:646543.696220}, \newcite{Riloff:1999:LDI:315149.315364}). Consequently, shifting to a new domain requires the user to not only name the target relations, but also to manually define new extraction rules or to annotate new training data by hand. Thus, those systems rely on extensive human involvement. In order to reduce the manual effort required by IE approaches, \newcite{Banko07} introduced a new extraction paradigm: Open IE. Unlike traditional IE methods, Open IE is not limited to a small set of target relations known in advance, but rather extracts all types of relations found in a text. In that way, it
facilitates the domain-independent discovery of relations extracted from text and scales to large, heterogeneous corpora such as the Web. Hence, \newcite{Banko07} identified three major challenges for Open IE systems:

\begin{description}
\item[Automation.] Open IE systems must \textit{rely on unsupervised extraction strategies}, i.e. instead of specifying target relations in advance, possible relations of interest must be automatically detected while making only a single pass over the corpus. Moreover, the manual labor of creating suitable training data or extraction patterns must be reduced to a minimum by requiring only a small set of hand-tagged seed instances or a few manually defined extraction patterns.
\item[Corpus Heterogeneity.] Heterogeneous datasets form an obstacle for profound linguistic tools such as syntactic or dependency parsers, since they commonly work well when trained and applied to a specific domain, but are prone to produce incorrect results when used in a different genre of text. Furthermore, Named Entity Recognition (NER) is unsuitable to target the variety and complexity of entity types on the Web. As Open IE systems are intended for \textit{domain-independent usage}, such tools should be avoided in favor of shallow parsing methods such as part-of-speech (POS) taggers.
\item[Efficiency.] In order to \textit{readily scale to large amounts of text}, Open IE systems must be computationally efficient. Enabling fast extraction over huge datasets, shallow linguistic features, like POS tags, are to be preferred over deep features that are derived from parse trees. 
\end{description}

These criteria were first implemented in the Open IE system \textsc{TextRunner}, which was presented together with the task definition in \newcite{Banko07}. This seminal work triggered a lot of research effort in this area, resulting in a multitude of proposed approaches that often did not strictly adhere to these initial guidelines. For example, to date, Open IE systems are commonly evaluated on rather small-scale, domain-dependent corpora. In addition, recent approaches frequently rely on the output of a dependency parser to identify extraction patterns, thereby hurting the domain-independence and efficiency assumptions.


\begin{figure}[!ht]
  \centering
    \tiny
    \begin{BVerbatim}[commandchars=\\\{\},codes={\catcode`$=3\catcode`_=8}]
\textbf{OLLIE:}
(1) (Republican candidate Mitt Romney; will be elected President in; 2008)[enabler=If he wins five key states]
(2) (Republican candidate Mitt Romney; will be elected; President)[enabler=If he wins five key states]
(3) (Mitt Romney; be candidate of; Republican)
(4) (Mitt Romney; be candidate for; Republican)
(5) (he; wins; five key states)

\textbf{ReVerb:}
(6) (he; wins; five key states)
(7) (Republican candidate Mitt Romney; will be elected President in; 2008)

\textbf{PredPatt:}
(8) (he; wins; five key states)
(9) (Republican candidate Mitt Romney; will be elected President in; 2008)

\textbf{ClausIE:}
(10) (he; wins; five key states)
(11) (Republican candidate Mitt Romney; will be elected; President in 2008 If he wins five key states)
(12) (Republican candidate Mitt Romney; will be elected; President in 2008)

\textbf{OpenIE 5.0:}
(13) (Republican candidate Mitt Romney; will be elected; President; T:in 2008)
(14) (he; wins; five key states)

\textbf{Graphene:}
(15) #1	CORE	(Mitt Romney; will be elected; President)
("a)	CONTEXT:NOUN\_BASED	Mitt Romney was a republican candidate .
("b)	CONTEXT:TEMPORAL	 in 2008 .
("c)	CONTEXT:CONDITION         #3
("d)	CONTEXT:NOUN\_BASED        #2
(16) #2	CORE	(Mitt Romney; was; a republican candidate)
(17) #3	CONTEXT	(he; wins; five key states)
    \end{BVerbatim}
  \caption{Comparison of the output generated by different Open IE systems for the input sentence \textit{"If he wins five key states, Republican candidate Mitt Romney will be elected President in 2008."}}
  \label{ComparativeAnalysis}
\end{figure}

\section{Open IE Systems}

A large body of work on the task of Open IE has been described since its introduction by \newcite{Banko07}. Existing Open IE approaches make use of a set of patterns in order to extract relational tuples from a sentence, each consisting of argument phrases and a phrase that expresses a semantic relation between them. Such extraction patterns are either hand-crafted or learned from automatically labeled training data, as shown below.




\subsection{Learning-based Systems}
The line of work on Open IE begins with \textsc{TextRunner} \cite{Banko07}, a self-supervised learning approach consisting of three modules. First, given a small sample of sentences from the Penn Treebank, the learner applies a dependency parser to heuristically identify and label a set of extractions as positive and negative training examples. This data is then used as input to a Naive Bayes classifier which learns a model of trustworthy relations using unlexicalized POS and noun phrase (NP) chunk features. The self-supervised nature mitigates the need for hand-labeled training data, and unlexicalized features help scale to the multitudes of relations found on the Web. The second component, the extractor, then generates candidate tuples by first identifying pairs of NP arguments and then heuristically designating each word in between as part of a relation phrase or not. Next, each candidate extraction is presented to the classifier, whereupon only those labeled as trustworthy are kept. Restricting to the use of shallow features in this step makes \textsc{TextRunner} highly efficient. Finally, a redundancy-based assessor assigns a probability to each retained tuple based on the number of sentences from which each extraction was found, thus exploiting the redundancy of information in Web text and assigning higher confidence to extractions that occur multiple times.

\begin{wrapfigure}{r}{10cm}
\centering
  \includegraphics[scale=0.28]{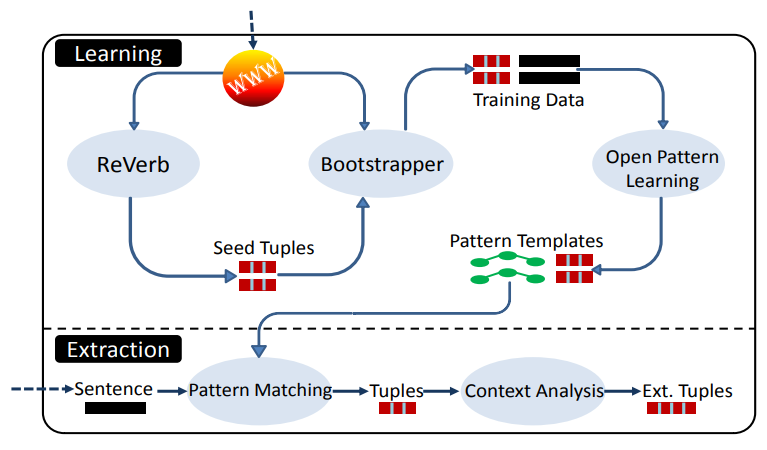}
  \caption{\textsc{OLLIE}'s system architecture \cite{Mausam12}. \textsc{OLLIE} begins with seed tuples from \textsc{ReVerb}, uses them to build a bootstrap learning set, and learns open pattern templates. These are applied to individual sentences at extraction time.}
  \label{fig:OllieArchitecture}
\end{wrapfigure}

\textsc{WOE} \cite{WuFei10} also learns an open information extractor without direct supervision. It makes use of Wikipedia as a source of training data by bootstrapping from entries in Wikipedia infoboxes, i.e. by heuristically matching infobox attribute-value pairs with corresponding sentences in the article. This data is then used to learn extraction patterns on both POS tags (\textsc{WOE}\textit{\textsuperscript{pos}}) and dependency parses (\textsc{WOE}\textit{\textsuperscript{parse}}). Former extractor utilizes a linear-chain Conditional Random Field (CRF) to train a model of relations on shallow features which outputs certain text between two NPs when it denotes a relation. Latter approach, in contrast, makes use of dependency trees to build a classifier that decides whether the shortest dependency path between two NPs indicates a semantic relation. By operating over dependency parses, even long-range dependencies can be captured. Accordingly, when comparing their two approaches, \newcite{WuFei10} show that the use of dependency features results in an increase in precision and recall over shallow linguistic features, though, at the cost of extraction speed, hence negatively affecting the scalability of the system.


\textsc{OLLIE} \cite{Mausam12} follows the idea of bootstrap learning of patterns based on dependency parse paths. However, while \textsc{WOE} relies on Wikipedia-based bootstrapping, \textsc{OLLIE} applies a set of high precision seed tuples from its predecessor system \textsc{ReVerb} (see section \ref{rule-based}) to bootstrap a large training set over which it learns a set of extraction pattern templates using dependency parses (see Figure~\ref{fig:OllieArchitecture}). In contrast to previously presented systems that fully ignore the context of a tuple and thus extract propositions that are not asserted as factual, but are only hypothetical or conditionally true, \textsc{OLLIE} includes a context-analysis step in which contextual information from the input sentence around an extraction is analyzed to expand the output representation by adding attribution and clausal modifiers, if necessary, and thus increasing the precision of the system (see extractions (1) and (2) in Figure~\ref{ComparativeAnalysis}; for details, see section \ref{nested}). Moreover, \textsc{OLLIE} is the first Open IE approach to identify not only verb-based relations, but also relationships mediated by nouns and adjectives (see extractions (3) and (4) in Figure~\ref{ComparativeAnalysis}). In that way, it expands the syntactic scope of relational phrases to cover a wider range of relation expressions, resulting in a much higher yield (at comparable precision) as compared to previous systems.


More recently, \newcite{Yahya2014} proposed ReNoun, an Open IE system that entirely focuses on the extraction of noun-mediated relations. Starting with a small set of high-precision seed facts relying on manually specified lexical patterns that are specifically tailored for NPs, a set of dependency parse patterns for the extraction of noun-based relations is learned with the help of distant supervision \cite{Mintz2009}. These patterns are then applied to generate a set of candidate extractions which are assigned a confidence score based on the frequency and coherence of the patterns producing them.






\subsection{Rule-based Systems}
\label{rule-based}

\begin{wrapfigure}{r}{7cm}
\centering
  \includegraphics[scale=0.12]{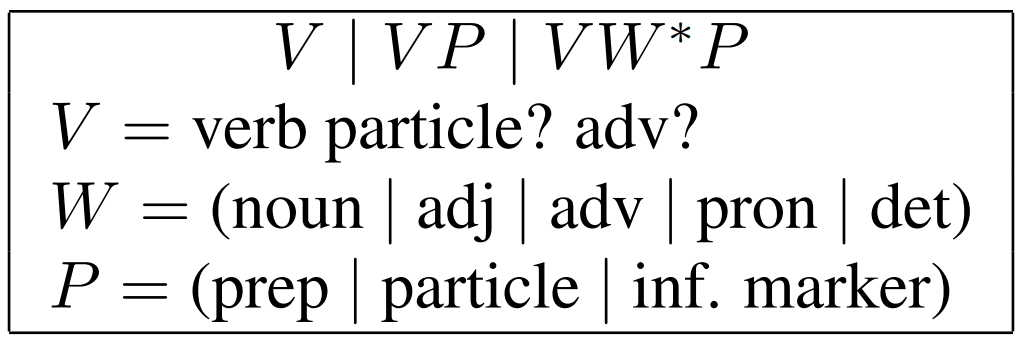}
  \caption{ReVerb's POS-based regular expression for reducing incoherent and uninformative extractions.}
  \label{fig:reverb}
\end{wrapfigure}

The second category of Open IE systems make use of hand-crafted extraction rules. This includes \textsc{ReVerb} \cite{Fader11}, a shallow extractor that addresses three common errors of hitherto existing Open IE systems: the output of such systems frequently contains a great many of uninformative extractions (i.e. extractions that omit critical information), incoherent extractions (i.e. extractions where the relational phrase has no meaningful interpretation) and overly-specific relations that convey too much information to be useful in further downstream semantic tasks. \textsc{ReVerb} improves over those approaches by introducing a syntactic constraint that is expressed in terms of a simple POS-based regular expression (see Figure~\ref{fig:reverb}), covering about 85\% of verb-based relational phrases in English text, as a linguistic analysis has revealed. In that way, the amount of incoherent and uninformative extractions is reduced. Moreover, in order to avoid overspecified relational phrases, a lexical constraint is presented which is based on the idea that a valid relational phrase should take many distinct arguments in a large corpus. Besides, while formerly proposed approaches start with the identification of candidate argument pairs, \textsc{ReVerb} follows a relation-centric approach by first determining relational phrases that satisfy above-mentioned constraints, and then finding a pair of NP arguments for each such phrase. An example output produced by ReVerb can be seen in Figure~\ref{ComparativeAnalysis} (6-7).

Whereas previously mentioned Open IE systems focus on the extraction of binary relations, commonly leading to extraction errors such as incomplete, uninformative or erroneous propositions, \textsc{KrakeN} \cite{Akbik12} is the first approach to be specifically built for capturing complete facts from sentences by gathering the full set of arguments for each relational phrase within a sentence, thus producing tuples of arbitrary arity. The identification of relational phrases and their corresponding arguments is based on hand-written extraction rules over typed dependency parses.

\textsc{Exemplar} \cite{Mesqu13} applies a similar approach for extracting n-ary relations, using hand-crafted patterns based on dependency parse trees to detect a relation trigger and the arguments connected to it. Based on the task of Semantic Role Labeling (SRL), whose key idea is to classify semantic constituents into different semantic roles \cite{Christensen10}, it assigns each argument its corresponding role (such as subject, direct object or prepositional object).

A more abstract approach, \textsc{PropS}, was suggested by \newcite{StanovskyFDG16}, who argue that it is hard to read out from a dependency parse the complete structure of a sentence's propositions, since, amongst others, different predications are represented in a non-uniform manner and proposition boundaries are not easy to detect. Therefore, they introduce a more semantically-oriented sentence representation that is generated by transforming a dependency parse tree into a directed graph which is tailored to directly represent the proposition structure of an input sentence. Consequently, extracting propositions from this novel output format is straightforward. The conversion of the dependency tree into the proposition structure is carried out by a rule-based converter.

PredPatt \cite{white2016universal} follows a similar approach. It employs a set of non-lexicalized rules defined over Universal Dependency (UD) parses \cite{DEMARNEFFE14.1062} to extract predicate-argument structures. In doing so, PredPatt constructs a directed graph, where a special dependency $ARG$ is built between the head token of a predicate and the head tokens of its arguments, while the original UD relations are preserved within predicate and argument phrases. As PredPatt uses language-agnostic patterns on UD structures, it is one of the few Open IE systems that work across different languages. 



\subsection{Clause-based Systems}
\label{paraphrase}
Aiming to improve the accuracy of Open IE approaches, more recent work is based on the idea of incorporating a sentence re-structuring stage whose goal is to transform complex sentences, where relations are spread over several clauses or presented in a non-canonical form, into a set of syntactically simplified independent clauses that are easy to segment into Open IE tuples. An example of such a paraphrase-based Open IE approach is ClausIE \cite{DelCorro13}, which exploits linguistic knowledge about the grammar of the English language to map the dependency relations of an input sentence to clause constituents. In that way, a set of coherent clauses presenting a simple linguistic structure is derived from the input. Then, the type of each clause is determined by combining knowledge of properties of verbs (with the help of domain-independent lexica) with knowledge about the structure of input clauses. Finally, based on its type, one or more propositions are generated from each clause, each representing different pieces of information. The basic set of patterns used for this task is shown in Table~\ref{fig:clausie}.

\begin{table}[ht]
\centering
\footnotesize
\begin{tabular}{c l l l l}
\hline
 & \textbf{Pattern} & \textbf{Clause type} & \textbf{Example} & \textbf{Derived clauses} \\ \hline
\textit{S\textsubscript{1}:} & SV\textsubscript{i} & SV & AE died. & (AE, died) \\
\textit{S\textsubscript{2}:} & SV\textsubscript{e}A & SVA & AE remained in Princeton. & (AE, remained, in Princeton) \\
\textit{S\textsubscript{3}:} & SV\textsubscript{c}C & SVC & AE is smart. & (AE, is, smart) \\
\textit{S\textsubscript{4}:} & SV\textsubscript{mt}O & SVO & AE has won the Nobel Prize. & (AE, has won, the Nobel Prize) \\
\textit{S\textsubscript{5}:} & SV\textsubscript{dt}O\textsubscript{i}O & SVOO & RSAS gave AE the Nobel Prize. & (RSAS, gave, AE, the Nobel Prize) \\
\textit{S\textsubscript{6}:} & SV\textsubscript{ct}OA & SVOA & The doorman showed AE to his office. & (The doorman, showed, AE, to his office) \\
\textit{S\textsubscript{7}:} & SV\textsubscript{ct}OC & SVOC & AE declared the meeting open. & (AE, declared, the meeting, open) \\ \hline
\end{tabular}
\caption{Basic patterns for proposition extraction \cite{DelCorro13}. \tiny{S: Subject, V: Verb, C: Complement, O: Direct object, A: Adverbial, V\textsubscript{i}: Intransitive verb, V\textsubscript{c}: Copular verb, V\textsubscript{e}: Extended-copular verb, V\textsubscript{mt}: Monotransitive verb, V\textsubscript{dt}: Ditransitive verb, V\textsubscript{ct}: Complex-transitive verb}}
  \label{fig:clausie}
\end{table}


In the same vein, \newcite{Schmid14} propose a strategy to break down structurally complex sentences into simpler ones by decomposing the original sentence into its basic building blocks via chunking. The dependencies of each two chunks are then determined (one of "connected", "disconnected" or "dependent") using either manually defined rules over dependency paths between words in different chunks or a Naive Bayes classifier trained on shallow features, such as POS tags and the distance between chunks. Depending on their relationships, chunks are combined into simplified sentences, upon which the extraction process is carried out.

\begin{figure}[ht]
\centering
  \includegraphics[scale=0.23]{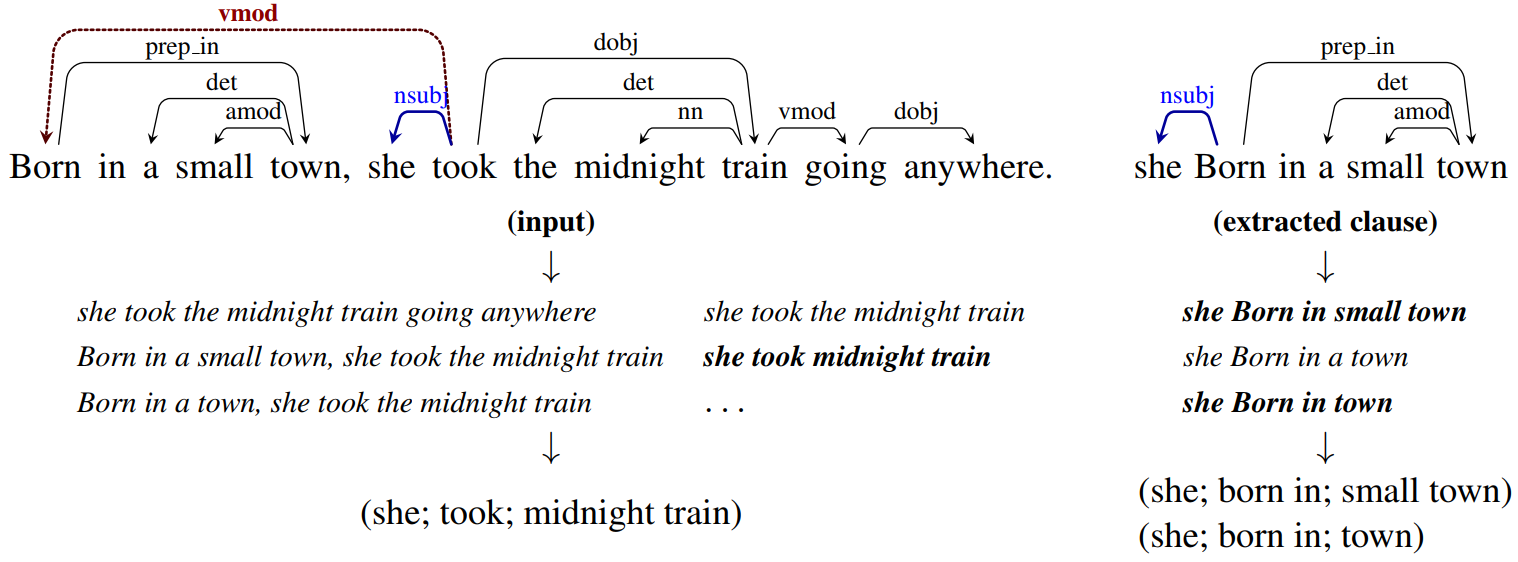}
  \caption{An illustration of Stanford Open IE's approach. From left to right, a sentence yields a number of independent clauses. From top to bottom, each clause produces a set of entailed shorter utterances, and segments the ones which match an atomic pattern into a relational triple \cite{Angeli15}.}
  \label{fig:stanford}
\end{figure}

\newcite{Angeli15} present Stanford Open IE, an approach in which a classifier is learned for splitting a sentence into a set of logically entailed shorter utterances by recursively traversing its dependency tree and predicting at each step whether an edge should yield an independent clause or not. In order to increase the usefulness of the extracted propositions for downstream applications, each self-contained clause is then maximally shortened by running natural logic inference over it. In the end, a small set of 14 hand-crafted patterns are used to extract a predicate-argument triple from each utterance. An illustration of this approach is depicted in Figure~\ref{fig:stanford}.

\subsection{Systems Capturing Inter-Proposition Relationships}
\label{nested}


Aforementioned Open IE systems lack the expressiveness needed for a proper interpretation of complex assertions, since they ignore the context under which a proposition is complete and correct. Thus, they do not distinguish between information asserted in a sentence and information that is only hypothetical or conditionally true. For example, extracting the relational tuple \texttuple{the earth}{is the center of}{the universe} from the sentence \textit{"Early scientists believed that the earth is the center of the universe."} would be inappropriate, since the input is not asserting it, but only noting that is was believed by early scientists \cite{Mausam16}. To properly handle such cases, \textsc{OLLIE} attempts a first solution by additionally extracting an attribution context, denoting a proposition that is reported or claimed by some entity: 
\begin{enumerate}
\item[] (\texttuple{the earth}{be the center of}{the university};\\
\texttt{AttributedTo} \textit{believe}; \textit{Early astronomers})
\end{enumerate}
In that way, it extends the default Open IE representation of $\langle arg_1, rel, arg_2 \rangle$ with an extra field. Besides, \textsc{OLLIE} pays attention to clausal modifiers, such as:
\begin{enumerate}
\item[] (\texttuple{Romney}{will be elected}{President};\\
\texttt{ClausalModifier} \textit{if}; \textit{he wins five key states})
\end{enumerate}
Both types of modifiers are identified by matching patterns with the dependency parse of the sentence. Clausal modifiers are determined by an adverbial-clause edge and filtered lexically (the first word of the clause must match a list of cue terms, e.g. \textit{if}, \textit{when}, or \textit{although}), while attribution modifiers are identified by a clausal-complement edge whose context verb must match one of the terms given in VerbNet's list of common verbs of communication and cognition \cite{Mausam12}.
A similar output is produced by \textsc{OLLIES}'s successor \textsc{OpenIE4} \cite{Mausam16}, which combines \textsc{SrlIE} \cite{Christensen10} and \textsc{RelNoun} \cite{Harinder16}. Former 
is a system that converts the output of a SRL system into an Open IE extraction by treating the verb as the relational phrase, while taking its role-labeled arguments as the Open IE argument phrases related to the relation. Latter, in contrast, represents a rule-based Open IE system that extracts noun-mediated relations, thereby paying special attention to demonyms and compound relational nouns. In addition, \textsc{OpenIE4} marks temporal and spatial arguments by assigning them a $T$ or $S$ label, respectively. Lately, its successor \textsc{OpenIE 5.0} was released\footnote{\url{https://github.com/dair-iitd/OpenIE-standalone}}. It integrates BONIE \cite{bonie2017} and OpenIEListExtractor\footnote{\url{https://github.com/swarnaHub/OpenIEListExtractor}}. While the former focuses on extracting tuples where one of the arguments is a number or a quantity-unit phrase, the latter targets the extraction of propositions from conjunctive sentences.

Similar to \textsc{OLLIE}, \newcite{bast2013open}, who explore the use of contextual sentence decomposition (CSD) for Open IE, advocate to further specify propositions with information on which they depend. Their system CSD-IE is based on the idea of paraphrasing-based approaches described in section \ref{paraphrase}. Using a set of hand-crafted rules over the output of a constituent parser, a sentence is first split into sub-sequences that semantically belong together, forming so-called "contexts". Each such context now contains a separate fact, yet it is often dependent on surrounding contexts. In order to preserve such inter-proposition relationships, tuples may contain references to other propositions. However, as opposed to \textsc{OLLIE}, where additional contextual modifiers are directly assigned to the corresponding relational tuples, \newcite{bast2013open} represent contextual information in the form of separate, linked propositions. To do so, each extraction is given a unique identifier that can be used in the argument position of an extraction for a later substitution with the corresponding fact by a downstream application. An example for an attribution is shown below \cite{bast2013open}:
\begin{enumerate}
\setlength{\itemsep}{0.001pt}
\item[\#1:] \texttuple{The Embassy}{said}{that \#2}
\item[\#2:] \texttuple{6,700 Americans}{were}{in Pakistan.}
\end{enumerate}
Another current approach that captures inter-proposition relationships is proposed by \newcite{bhutani2016nested}, who present a nested representation for Open IE that is able to capture high-level dependencies, allowing for a more accurate representation of the meaning of an input sentence. Their system \textsc{NestIE} uses bootstrapping over a dataset for textual entailment to learn both binary and nested triple representations for n-ary relations over dependency parse trees. These patterns can take on the form of binary triples \tuple{$arg_1$}{$rel$}{$arg_2$} or nested triples such as \tuple{\tuple{$arg_1$}{$rel$}{$arg_2$}}{$rel_2$}{$arg_3$} for n-ary relations. Using a set of manually defined rules, contextual links between extracted propositions are inferred from the dependency parse in order to generate a nested representation of assertions that are complete and closer in meaning to the original statement. Similar to \textsc{OLLIE}, contextual links are identified as clausal complements, conditionals and relation modifiers. Linked propositions are represented by arguments that refer to the corresponding propositions using identifiers, e.g. \cite{bhutani2016nested}:

\begin{enumerate}
\setlength{\itemsep}{0.001pt}
\item[\#1:] \tuple{\textit{body}}{\textit{appeared to have been thrown}}{$\emptyset$}
\item[\#2:] \tuple{\#1}{\textit{from}}{\textit{vehicle}}
\end{enumerate}

or another example based on the following sentence:

\begin{enumerate}
\item[] \inlinetext{After giving 5,000 people a second chance at life, doctors are celebrating the 25th anniversary of Britain's first heart transplant.}
\end{enumerate}

\begin{enumerate}
\setlength{\itemsep}{0.001pt}
\item[\#1:] \texttuple{doctors}{are celebrating}{the 25th anniversary of Britain's first heart transplant}
\item[\#2:] \texttuple{doctors}{giving}{second chance at life}
\item[\#3:] \tuple{\#1}{\textit{after}}{\#2}
\end{enumerate}


MinIE \cite{gashteovski17}, another recent Open IE system, is built on top of ClausIE, a system that was found to often produce overly specific extractions. Such overly specific constituents that combine multiple, potentially semantically unrelated propositions in a single relational or argument phrase generally hurt the performance of downstream semantic applications, such as question answering or textual entailment. In fact, those approaches benefit from extractions that are as compact as possible. Therefore, MinIE aims to minimize both relational and argument phrases by identifying and removing parts that are considered overly specific. For this purpose, MinIE provides four different minimization modes which differ in their aggressiveness, thus allowing to control the trade-off between precision and recall. Moreover, it semantically annotates extractions with information about polarity, modality, attribution and quantities instead of directly representing it in the actual extractions, as the following example shows \cite{gashteovski17}:

\begin{enumerate}
\item[] \inlinetext{Pinocchio believes that the hero Superman was not actually born on beautiful Krypton.}
\end{enumerate}

\begin{enumerate}
\setlength{\itemsep}{0.001pt}
\item[\#1:] \texttuple{Superman}{was born actually on}{beautiful Krypton}\\
\textit{Annotation: factuality, (- [not], certainty), attribution (Pinocchio, +, possibility [believes])}
\item[\#2:] \texttuple{Superman}{was born on}{beautiful Krypton}\\
\textit{Annotation: factuality, (- [not], certainty), attribution (Pinocchio, +, possibility [believes])}
\item[\#3:] \texttuple{Superman}{"is"}{hero}\\
\textit{Annotation: factuality, (+, certainty)}
\item[] with + and - signifying positive and negative polarity, respectively.
\end{enumerate}

In that way, the output generated by MinIE is further reduced to its core constituents, producing maximally shortened, semantically enriched extractions.

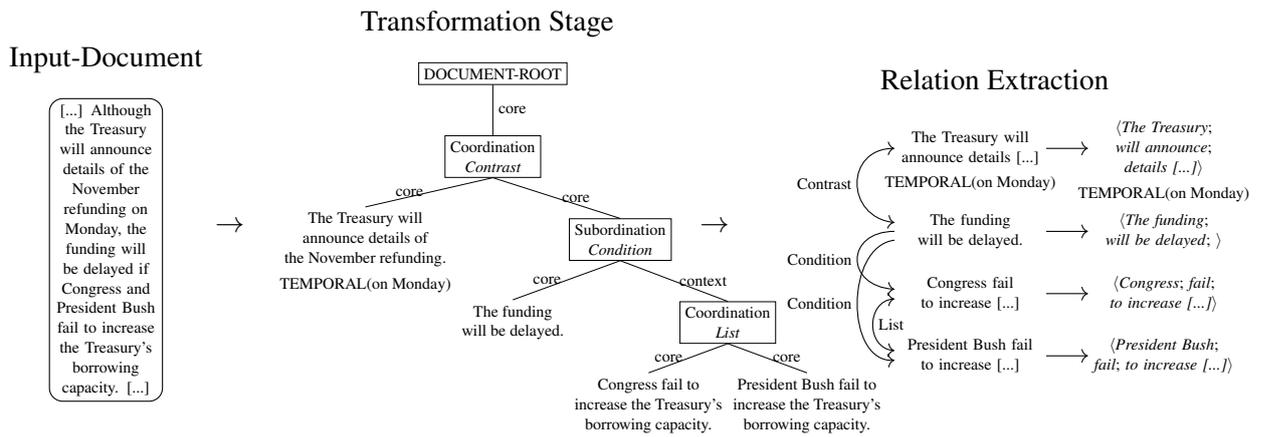
\begin{figure}[ht]
\centering

\begin{minipage}{0.18\textwidth}
\centering
Input-Document\\
\vspace{0.3cm}
\begin{tikzpicture}[scale=0.55, every node/.style={align=center, transform shape}]
\node[rectangle, solid, draw=black, text width=0.85*\columnwidth, rounded corners=5pt]{
[...] Although the Treasury will announce details of the November refunding on Monday, the funding will be delayed if Congress and President Bush fail to increase the Treasury's borrowing capacity. [...]
};
\end{tikzpicture}
\end{minipage}%
$\rightarrow$
\hspace{0.2cm}
\begin{minipage}{0.35\textwidth}
\centering
Transformation Stage\\
\vspace{0.3cm}
\begin{tikzpicture}[scale=0.55, level distance=2cm, sibling distance=0cm, every tree node/.style={align=center, transform shape}]
\Tree [
.\node[style={draw,rectangle}] {DOCUMENT-ROOT}; 
  \edge node[midway, right] {core}; [
      .\node [style={draw,rectangle}] {Coordination\\\textit{Contrast}};
            \edge node[midway, left] {core}; [.\node(a)[label=below:TEMPORAL(on Monday)]{The Treasury will\\announce details of\\the November refunding.};]
            \edge node[midway, right] {core}; [
              .\node [style={draw,rectangle}] {Subordination\\\textit{Condition}};
                \edge node[midway, left] {core}; [.\node(b){The funding\\will be delayed.};]
                \edge node[midway, right] {context}; [
                  .\node [style={draw,rectangle}] {Coordination\\\textit{List}};
                    \edge node[midway, left] {core}; [.\node(c){Congress fail to\\increase the Treasury's\\borrowing capacity.};]
                    \edge node[midway, right] {core}; [.\node(d){President Bush fail to\\increase the Treasury's\\borrowing capacity.};]
                ]
            ]
        ]
    ]
]
\end{tikzpicture}
\end{minipage}%
$\rightarrow$
\hspace{0.5cm}
\begin{minipage}{0.35\textwidth}
\centering
Relation Extraction\\
\vspace{0.3cm}
\begin{tikzpicture}[scale=0.55, every node/.style={align=center, transform shape}]
    \node(A)[text width=0.6*\linewidth, label=below:TEMPORAL(on Monday)]{The Treasury will announce details [...]};
    \node(B)[below of=A, yshift=-1cm, text width=0.6*\linewidth]{The funding will be delayed.};
    \node(C)[below of=B, yshift=-0.5cm, text width=0.6*\linewidth]{Congress fail to increase [...]};
    \node(D)[below of=C, yshift=-0.5cm, text width=0.6*\linewidth]{President Bush fail to increase [...]};

    \node(a)[right=1 of A, text width=0.6*\linewidth,  label=below:TEMPORAL(on Monday)]{\texttuple{The Treasury}{will announce}{details [...]}};
    \node(b)[right=1 of B, text width=0.6*\linewidth]{\texttuple{The funding}{will be delayed}{}};
    \node(c)[right=1 of C, text width=0.6*\linewidth]{\texttuple{Congress}{fail}{to increase [...]}};
    \node(d)[right=1 of D, text width=0.6*\linewidth]{\texttuple{President Bush}{fail}{to increase [...]}};

    \draw[solid, <->] (A)..controls +(west:3) and +(west:3)..([yshift=10]B) node [left, midway] () {Contrast};
    \draw[solid, ->] (B)..controls +(west:3) and +(west:3)..([yshift=5]C) node [left, midway] () {Condition};
    \draw[solid, ->] ([yshift=-10]B)..controls +(west:3) and +(west:3)..([yshift=-5]D) node [left, midway] () {Condition};
    \draw[solid, <->] ([yshift=-5]C)..controls +(west:2.5) and +(west:2.5)..([yshift=5]D) node [right, midway] () {List};

    \draw[solid, ->] (A) to (a);
    \draw[solid, ->] (B) to (b);
    \draw[solid, ->] (C) to (c);
    \draw[solid, ->] (D) to (d);

\end{tikzpicture}\\
\end{minipage}

\caption{Graphene's extraction workflow for an example sentence \cite{Cetto2018}.}
\label{fig:workflow}
\end{figure}

To further enhance the expressiveness of extracted propositions and sustain their interpretability in downstream artificial intelligence tasks, \newcite{Cetto2018} propose Graphene, an Open IE framework that uses a set of hand-crafted simplification rules to transform complex natural language sentences into clean, compact structures by removing clauses and phrases that present no central information from the input and converting them into stand-alone sentences. In that way, a source sentence is transformed into a hierarchical representation in the form of core facts and accompanying contexts \cite{Niklaus2016}. In addition, inspired by the work on Rhetorical Structure Theory \cite{mann1988rhetorical}, a set of syntactic and lexical patterns is used to identify the rhetorical relations by which core sentences and their associated contexts are connected in order to preserve their semantic relationships and return a set of semantically typed and interconnected relational tuples (see extractions (15-17) in Figure~\ref{ComparativeAnalysis}). Graphene's extraction workflow is illustrated in Figure~\ref{fig:workflow}.

\section{Evaluation}
\label{sec:eval}



Though a multitude of systems for Open IE have been developed over the last decade, a clear formal specification of what constitutes a valid relational tuple is still missing. This lack of a well-defined, generally accepted task definition prevented the creation of an established, large-scale annotated corpus serving as a gold standard dataset for an objective and reproducible cross-system comparison. As a consequence, to date, Open IE systems were predominantly evaluated by hand on small-scale corpora that consist of only a few hundred sentences, thereby ignoring one of the fundamental goals of Open IE: scalability to large amounts of text. Moreover, none of the datasets that were used for assessing the performance of different systems is widely agreed upon. As can be seen in Table \ref{evaluation}, the corpora compiled by \newcite{DelCorro13}, \newcite{Xu13}, \newcite{Fader11} and \newcite{Banko07} are occasionally re-used. However, new datasets are still collected, hindering a fair comparison of the proposed approaches. Besides, although Open IE methods are targeted at being domain independent and able to cope with heterogeneous datasets, the corpora used in the evaluation process are restricted to the news, Wikipedia and Web domains for the most part. Accordingly, no clear statement about the portability of the approaches to various genres of text is possible. In addition, most evaluation procedures described in the literature focus on precision-oriented metrics, while either completely ignoring recall or using some kind of proxy, such as yield, i.e. the total number of extractions labeled as correct, or coverage, i.e. the percentage of text from the input that is contained in at least one of the extractions. Hence, the absence of a standard evaluation procedure makes it hard to replicate and compare the performance of different Open IE systems. Table \ref{evaluation} provides a detailed overview of both the datasets and measures used for intrinsically evaluating the various approaches described above, while Table \ref{evaluationextrinsic} shows the tasks that were used for an extrinsic evaluation of a small set of Open IE systems.


\begin{table}
\begin{tcolorbox}[tab2,tabularx={|v{1.5cm}||v{3.2cm}|v{3.5cm}|Z|}, fontupper=\scriptsize]
\textbf{System} & \textbf{Baselines} & \textbf{\# sentences and domain} & \textbf{Metrics} \\ \hline \hline
\textsc{TextRunner} & \textsc{KnowItAll} \cite{Etzioni04} & 400 Web & \% correct extractions  \\ \hline
\textsc{WOE} & \textsc{TextRunner} & 300 news\\ 300 Wikipedia \\ 300 Web & precision-recall curve  \\ \hline
\textsc{OLLIE} & \textsc{ReVerb}\\ \textsc{WOE}\textit{\textsuperscript{parse}} & 300 news (from \textsc{WOE})\\300 Wikipedia (from \textsc{WOE})\\300 biology & precision-yield curve \\ \hline

\textsc{ReVerb} & \textsc{TextRunner} \\ \textsc{WOE}\textit{\textsuperscript{pos}} \\ \textsc{WOE}\textit{\textsuperscript{parse}} & 500 Web & precision-recall curve  \\ \hline

\textsc{KrakeN} & \textsc{ReVerb} & 500 Web (from \textsc{ReVerb}) & \begin{compactitem}
\item precision
\item completeness
\item \# facts extracted per sentence
\end{compactitem}  \\ \hline
\textsc{Exemplar} & \textsc{ReVerb}\\ \textsc{OLLIE} \\ \textsc{Sonex} \cite{Merhav12} \\ \textsc{Patty} \cite{Nakashole:2012:PTR:2390948.2391076} \\ \textsc{TreeKernel} \cite{Xu13} \\ \textsc{SwiRL} \cite{surdeanu2003using}\\ \textsc{Lund} \cite{johansson2008dependency}  & 500 Web (from \textsc{TextRunner})\\500 news\\100 news (from \textsc{TreeKernel})\\222 news &
$\ast$ binary:
\begin{compactitem}
\item precision
\item recall
\item F\textsubscript{1}-score
\end{compactitem}
$\ast$ n-ary:
\begin{compactitem}
\item precision over arguments
\item recall over arguments
\end{compactitem} \\ \hline

PredPatt \\ \cite{zhang-EtAl:2017:IWCS} & \textsc{OLLIE}\\ ClausIE \\ Stanford Open IE\\ \textsc{OpenIE4} & 13k Web \\36k news & precision-recall curve  \\ \hline
ClausIE & \textsc{TextRunner}\\\textsc{WOE}\textit{\textsuperscript{parse}}\\\textsc{ReVerb}\\\textsc{OLLIE}\\\textsc{KrakeN}& 500 Web (from \textsc{ReVerb})\\200 Wikipedia \\200 news &
\begin{compactitem}
\item precision-yield curve
\item \% correct extractions
\end{compactitem} \\ \hline

\newcite{Schmid14} & \textsc{ReVerb} \\ \textsc{Exemplar} & 500 Web (from \textsc{TextRunner})\\500 news\\100 news (from \textsc{TreeKernel}) & 
\begin{compactitem}
\item precision
\item recall
\item F\textsubscript{1}-score
\item time per sentence before and after sentence re-structuring
\end{compactitem}  \\ \hline
\textsc{OpenIE4} & \textsc{ReVerb}\\ \textsc{OLLIE} & not reported & \begin{compactitem}
\item precision
\item yield
\end{compactitem} \\ \hline
\textsc{CSD-IE} & \textsc{ReVerb}\\\textsc{OLLIE}\\ClausIE & 200 Wikipedia (from ClausIE)\\200 news (from ClausIE) &
\begin{compactitem}
\item \% triples labeled accurate
\item \% correct triples labeled minimal
\item coverage (\% text contained in at least one triple)
\item average triple length
\end{compactitem}
 \\ \hline

\textsc{NestIE} &\textsc{ReVerb}\\\textsc{OLLIE}\\ ClausIE & 200 Wikipedia (from ClausIE) \\ 200 news (from ClausIE) &
\begin{compactitem}
\item correctness (0/1)
\item minimality (0/1)
\item informativeness (0-5)
\end{compactitem} \\ \hline

\textsc{MinIE} & ClausIE\\ \textsc{OLLIE}\\ Stanford Open IE & 10k news (from \newcite{Sandhaus08})\\200 news (from ClausIE)\\200 Wikipedia (from ClausIE) &
\begin{compactitem}
\item \# extractions
\item \# non-redundant extractions
\item recall
\item factual precision
\item attribution precision
\item mean word count per triple (proxy for minimality)
\end{compactitem}

\\\hline

Graphene &\textsc{OLLIE}\\\textsc{ReVerb}\\\textsc{PropS}\\ClausIE\\Stanford Open IE\\\textsc{OpenIE4} & 3,200 Wikipedia and news \cite{Stanovsky2016EMNLP} & precision-recall curve \\ \hline

\end{tcolorbox}
\caption{Comparison of the intrinsic evaluation approaches applied by the different Open IE systems.}
\label{evaluation}
\end{table}

\begin{table}
\begin{tcolorbox}[tab2,tabularx={|v{3cm}||Z|}, fontupper=\footnotesize]
\textbf{System} & \textbf{Task} \\ \hline \hline
\textsc{PropS} & MCTest comprehension task \cite{richardson2013mctest}\\\hline
Stanford Open IE & TAC KBP Slot Filling Challenge \cite{surdeanu2013overview} \\\hline
\end{tcolorbox}
\caption{Extrinsic evaluation approaches.}
\label{evaluationextrinsic}
\end{table}

In order to address aforementioned difficulties, \newcite{Stanovsky2016EMNLP} recently made a first attempt in standardizing the Open IE evaluation by providing a large gold benchmark corpus. It is based on a set of consensual guiding principles that underly most Open IE approaches proposed so far, as they have identified. Those principles cover the core aspects of the task of Open IE, allowing for a clearer formulation of the problem to be solved. The three key features to consider are the following:

\begin{description}
\item[Assertedness.] The assertedness principle states that extracted propositions should be asserted by the original sentence. Usually, instead of inferring propositions out of implied statements, e.g. the tuple \texttuple{Sam}{convinced}{John} out of \texttuple{Sam}{succeeded in convincing}{John}, Open IE systems tend to extract the full relational phrase (\texttuple{Sam}{succeeded in convincing}{John}), incorporating matrix verbs (\inlinetext{succeeded}) and other elements, such as negotiations or modals (e.g. \texttuple{John}{could not join}{the band}).

\item[Minimal Propositions.] In order to serve for semantic tasks, it is beneficial for Open IE systems to extract compact, self-contained propositions that do not combine several unrelated facts. Therefore, systems should aim to generate valid propositions with minimal spans for both relation and argument slots, while preserving the meaning of the input. As an example, the coordination in the sentence \inlinetext{Bell distributes electronic and building products} should ideally yield the two propositions: \texttuple{Bell}{distributes}{electronic products} and \texttuple{Bell}{distributes}{building products}.

\item[Completeness and Open Lexicon.] The completeness and open lexicon principle aims to extract all relations that are asserted in the input text. 
This principle was one of the fundamental ideas that have been introduced in the work of \newcite{Banko07} together with the Open IE terminology. In their work, the Open IE task was defined as a domain-independent task which extracts all possible relations from heterogeneous corpora, instead of only extracting a set of pre-specified classes of relations. The majority of current Open IE systems realize this challenge by considering all possible verbs as potential relations. Accordingly, their scope is often limited to the extraction of verbal predicates, while ignoring relations mediated by more complex syntactic constructs, such as nouns or adjectives.
\end{description}

Realizing that above-mentioned requirements are subsumed by the task of Question Answering (QA) driven Semantic Role Labeling (SRL) \cite{he2015question}, \newcite{Stanovsky2016EMNLP} converted the annotations of a QA-SRL dataset to an Open IE corpus, resulting in more than 10,000 extractions over 3,200 sentences from Wikipedia and the Wall Street Journal.

In addition, \newcite{Schneider17} presented RelVis, another benchmark framework for Open IE that allows for a large-scale comparative analysis of Open IE approaches. Besides \newcite{Stanovsky2016EMNLP}'s benchmark suite, it comprises the n-ary news dataset proposed in \newcite{Mesqu13}, \newcite{Banko07}'s Web corpus and the Penn sentences from \newcite{Xu13}. Similar to the toolkit proposed in \newcite{Stanovsky2016EMNLP}, RelVis supports a quantitative evaluation of the performance of Open IE systems in terms of precision, recall and F\textsubscript{2}-score. In addition, it facilitates a manual qualitative error analysis. For this purpose, six common error classes are distinguished to which inaccurate extractions can be assigned: (1) wrong boundaries, where the relational or argument phrase is either too long or too small; (2) redundant extraction, where the proposition asserted in an extraction is already expressed in another extraction; (3) uninformative extraction, where critical information is omitted; (4) missing extraction, i.e. a false negative, where either a relation is not detected by the system or the argument-finding heuristics choose the wrong arguments or none argument at all; (5)  wrong extraction, where no meaningful interpretation of the proposition is possible; and (6) out of scope extraction, where a system yields a correct extraction that was not recognized by the authors of the gold dataset.

\section{Open Research Questions}

Even after more than one decade of research in the area of Open IE, there are still a lot of open research questions. As shown in Section \ref{sec:eval}, to date, there is only very little work on evaluating and comparing results among different Open IE systems in a large-scale, objective and reproducible fashion. Instead, most approaches use proprietary datasets over small, domain-dependent corpora. \newcite{Stanovsky2016EMNLP} and \newcite{Schneider17} recently made the first move to standardize the evaluation of Open IE by proposing benchmark frameworks that operate on a larger scale. However, apart from \newcite{Cetto2018}, neither benchmarking toolkit has been adopted yet in the Open IE community. 

Moreover, most Open IE approaches focus on the English language, leaving aside other languages. Notable exceptions are \newcite{Falke2016}, who investigate the portability of the \textsc{PropS} system to German, and \newcite{Gamallo15multilingual}, who propose a multilingual system covering English, Spanish, Portuguese and Galician. Besides, due to the use of language-agnostic patterns over UD parses, PredPatt works across languages, though its performance was only evaluated on English texts. Hence, the applicability and transferability of previously proposed Open IE approaches to other languages than English represents an interesting direction for future work.

Finally, the problem of canonicalizing relational phrases and arguments has been hardly addressed so far. However, normalizing extractions would be highly beneficial for downstream semantic tasks, such as textual entailment or knowledge base population. Besides, coreference resolution, another field that has largely been ignored to date, may assist those tasks with the interpretation of the extracted propositions.

\section{Conclusion}
We presented an overview of the various methods that were proposed to solve the task of Open IE. We classified them into the four categories of learning-based, rule-based, clause-based systems and approaches capturing inter-proposition relationships, thereby showing their evolution over time, as well as the specific problems they tackle. Moreover, we described the approaches that were used to assess the performance of the proposed Open IE systems, while depicting the gaps in the evaluation procedures that are commonly applied to date. Finally, we identified directions for future work.

\bibliographystyle{acl}
\bibliography{graphene}

\begin{thebibliography}{}

\bibitem[\protect\citename{Agichtein and
  Gravano}2000]{Agichtein:2000:SER:336597.336644}
Eugene Agichtein and Luis Gravano.
\newblock 2000.
\newblock Snowball: Extracting relations from large plain-text collections.
\newblock In {\em Proceedings of the Fifth ACM Conference on Digital
  Libraries}, DL '00, pages 85--94, New York, NY, USA. ACM.

\bibitem[\protect\citename{Akbik and L{\"o}ser}2012]{Akbik12}
Alan Akbik and Alexander L{\"o}ser, 2012.
\newblock {\em Proceedings of the Joint Workshop on Automatic Knowledge Base
  Construction and Web-scale Knowledge Extraction (AKBC-WEKEX)}, chapter
  KrakeN: N-ary Facts in Open Information Extraction, pages 52--56.
\newblock Association for Computational Linguistics.

\bibitem[\protect\citename{Angeli \bgroup et al.\egroup }2015]{Angeli15}
Gabor Angeli, Melvin~Jose Johnson~Premkumar, and Christopher~D. Manning.
\newblock 2015.
\newblock Leveraging linguistic structure for open domain information
  extraction.
\newblock In {\em Proceedings of the 53rd Annual Meeting of the Association for
  Computational Linguistics and the 7th International Joint Conference on
  Natural Language Processing (Volume 1: Long Papers)}, pages 344--354,
  Beijing, China, July. Association for Computational Linguistics.

\bibitem[\protect\citename{Banko \bgroup et al.\egroup }2007]{Banko07}
Michele Banko, Michael~J. Cafarella, Stephen Soderland, Matt Broadhead, and
  Oren Etzioni.
\newblock 2007.
\newblock Open information extraction from the web.
\newblock In {\em Proceedings of the 20th International Joint Conference on
  Artifical Intelligence}, pages 2670--2676, San Francisco, CA, USA. Morgan
  Kaufmann Publishers Inc.

\bibitem[\protect\citename{Bast and Haussmann}2013]{bast2013open}
Hannah Bast and Elmar Haussmann.
\newblock 2013.
\newblock Open information extraction via contextual sentence decomposition.
\newblock In {\em Semantic Computing (ICSC), 2013 IEEE Seventh International
  Conference on}, pages 154--159. IEEE.

\bibitem[\protect\citename{Bhutani \bgroup et al.\egroup
  }2016]{bhutani2016nested}
Nikita Bhutani, H.~V. Jagadish, and Dragomir~R. Radev.
\newblock 2016.
\newblock Nested propositions in open information extraction.
\newblock In {\em Proceedings of the 2016 Conference on Empirical Methods in
  Natural Language Processing, {EMNLP} 2016, Austin, Texas, USA, November 1-4,
  2016}, pages 55--64.

\bibitem[\protect\citename{Brin}1999]{Brin:1998:EPR:646543.696220}
Sergey Brin.
\newblock 1999.
\newblock Extracting patterns and relations from the world wide web.
\newblock In {\em Selected Papers from the International Workshop on The World
  Wide Web and Databases}, WebDB '98, pages 172--183, London, UK, UK.
  Springer-Verlag.

\bibitem[\protect\citename{Cetto \bgroup et al.\egroup }2018]{Cetto2018}
Matthias Cetto, Christina Niklaus, Andr\'{e} Freitas, and Siegfried Handschuh.
\newblock 2018.
\newblock Graphene: Semantically-linked propositions in open information
  extraction.
\newblock In {\em Prooceedings of COLING 2018. To appear.}

\bibitem[\protect\citename{Christensen \bgroup et al.\egroup
  }2010]{Christensen10}
Janara Christensen, Mausam, Stephen Soderland, and Oren Etzioni.
\newblock 2010.
\newblock Semantic role labeling for open information extraction.
\newblock In {\em Proceedings of the NAACL HLT 2010 First International
  Workshop on Formalisms and Methodology for Learning by Reading}, FAM-LbR '10,
  pages 52--60, Stroudsburg, PA, USA. Association for Computational
  Linguistics.

\bibitem[\protect\citename{Del~Corro and Gemulla}2013]{DelCorro13}
Luciano Del~Corro and Rainer Gemulla.
\newblock 2013.
\newblock Clausie: Clause-based open information extraction.
\newblock In {\em Proceedings of the 22Nd International Conference on World
  Wide Web}, pages 355--366, New York, NY, USA. ACM.

\bibitem[\protect\citename{Etzioni \bgroup et al.\egroup }2004]{Etzioni04}
Oren Etzioni, Michael Cafarella, Doug Downey, Stanley Kok, Ana-Maria Popescu,
  Tal Shaked, Stephen Soderland, Daniel~S. Weld, and Alexander Yates.
\newblock 2004.
\newblock Web-scale information extraction in knowitall: (preliminary results).
\newblock In {\em Proceedings of the 13th International Conference on World
  Wide Web}, WWW '04, pages 100--110, New York, NY, USA. ACM.

\bibitem[\protect\citename{Fader \bgroup et al.\egroup }2011]{Fader11}
Anthony Fader, Stephen Soderland, and Oren Etzioni.
\newblock 2011.
\newblock Identifying relations for open information extraction.
\newblock In {\em Proceedings of the 2011 Conference on Empirical Methods in
  Natural Language Processing}, pages 1535--1545, Edinburgh, Scotland, UK.,
  July. Association for Computational Linguistics.

\bibitem[\protect\citename{Falke \bgroup et al.\egroup }2016]{Falke2016}
Tobias Falke, Gabriel Stanovsky, Iryna Gurevych, and Ido Dagan.
\newblock 2016.
\newblock Porting an open information extraction system from english to german.
\newblock In {\em Proceedings of the 2016 Conference on Empirical Methods in
  Natural Language Processing (EMNLP)}, pages 892--898. Association for
  Computational Linguistics, November.

\bibitem[\protect\citename{Gamallo and Garcia}2015]{Gamallo15multilingual}
Pablo Gamallo and Marcos Garcia.
\newblock 2015.
\newblock Multilingual open information extraction.
\newblock In Francisco Pereira, Penousal Machado, Ernesto Costa, and
  Am{\'i}lcar Cardoso, editors, {\em Progress in Artificial Intelligence},
  pages 711--722, Cham. Springer International Publishing.

\bibitem[\protect\citename{Gashteovski \bgroup et al.\egroup
  }2017]{gashteovski17}
Kiril Gashteovski, Rainer Gemulla, and Luciano del Corro.
\newblock 2017.
\newblock Minie: minimizing facts in open information extraction.
\newblock In {\em The Conference on Empirical Methods in Natural Language
  Processing - proceedings of System Demonstrations : September 9-11, 2017,
  Copenhagen, Denmark : EMNLP 2017}, pages 2620--2630, Stroudsburg, PA.
  Association for Computational Linguistics.

\bibitem[\protect\citename{He \bgroup et al.\egroup }2015]{he2015question}
Luheng He, Mike Lewis, and Luke Zettlemoyer.
\newblock 2015.
\newblock Question-answer driven semantic role labeling: Using natural language
  to annotate natural language.
\newblock In {\em Proceedings of the 2015 conference on empirical methods in
  natural language processing}, pages 643--653.

\bibitem[\protect\citename{Johansson and Nugues}2008]{johansson2008dependency}
Richard Johansson and Pierre Nugues.
\newblock 2008.
\newblock Dependency-based semantic role labeling of propbank.
\newblock In {\em Proceedings of the Conference on Empirical Methods in Natural
  Language Processing}, pages 69--78. Association for Computational
  Linguistics.

\bibitem[\protect\citename{Jurafsky and Martin}2009]{Jurafsky:2009:SLP:1214993}
Daniel Jurafsky and James~H. Martin.
\newblock 2009.
\newblock {\em Speech and Language Processing (2Nd Edition)}.
\newblock Prentice-Hall, Inc., Upper Saddle River, NJ, USA.

\bibitem[\protect\citename{Mann and Thompson}1988]{mann1988rhetorical}
William~C Mann and Sandra~A Thompson.
\newblock 1988.
\newblock Rhetorical structure theory: Toward a functional theory of text
  organization.
\newblock {\em Text-Interdisciplinary Journal for the Study of Discourse},
  8(3):243--281.

\bibitem[\protect\citename{Marneffe \bgroup et al.\egroup
  }2014]{DEMARNEFFE14.1062}
Marie-Catherine~De Marneffe, Timothy Dozat, Natalia Silveira, Katri Haverinen,
  Filip Ginter, Joakim Nivre, and Christopher~D. Manning.
\newblock 2014.
\newblock Universal stanford dependencies: a cross-linguistic typology.
\newblock In Nicoletta Calzolari~(Conference Chair), Khalid Choukri, Thierry
  Declerck, Hrafn Loftsson, Bente Maegaard, Joseph Mariani, Asuncion Moreno,
  Jan Odijk, and Stelios Piperidis, editors, {\em Proceedings of the Ninth
  International Conference on Language Resources and Evaluation (LREC'14)},
  Reykjavik, Iceland, may. European Language Resources Association (ELRA).

\bibitem[\protect\citename{Mausam \bgroup et al.\egroup }2012]{Mausam12}
Mausam, Michael Schmitz, Stephen Soderland, Robert Bart, and Oren Etzioni.
\newblock 2012.
\newblock Open language learning for information extraction.
\newblock In {\em Proceedings of the 2012 Joint Conference on Empirical Methods
  in Natural Language Processing and Computational Natural Language Learning},
  pages 523--534, Jeju Island, Korea, July. Association for Computational
  Linguistics.

\bibitem[\protect\citename{Mausam}2016]{Mausam16}
Mausam.
\newblock 2016.
\newblock Open information extraction systems and downstream applications.
\newblock In {\em Proceedings of the Twenty-Fifth International Joint
  Conference on Artificial Intelligence, {IJCAI} 2016, New York, NY, USA, 9-15
  July 2016}, pages 4074--4077.

\bibitem[\protect\citename{Merhav \bgroup et al.\egroup }2012]{Merhav12}
Yuval Merhav, Filipe Mesquita, Denilson Barbosa, Wai~Gen Yee, and Ophir
  Frieder.
\newblock 2012.
\newblock Extracting information networks from the blogosphere.
\newblock {\em ACM Transactions on the Web}, 6(3):11:1--11:33, September.

\bibitem[\protect\citename{Mesquita \bgroup et al.\egroup }2013]{Mesqu13}
Filipe Mesquita, Jordan Schmidek, and Denilson Barbosa.
\newblock 2013.
\newblock Effectiveness and efficiency of open relation extraction.
\newblock In {\em Proceedings of the 2013 Conference on Empirical Methods in
  Natural Language Processing}, pages 447--457. Association for Computational
  Linguistics.

\bibitem[\protect\citename{Mintz \bgroup et al.\egroup }2009]{Mintz2009}
Mike Mintz, Steven Bills, Rion Snow, and Daniel Jurafsky.
\newblock 2009.
\newblock Distant supervision for relation extraction without labeled data.
\newblock In {\em Proceedings of the Joint Conference of the 47th Annual
  Meeting of the ACL and the 4th International Joint Conference on Natural
  Language Processing of the AFNLP}, pages 1003--1011. Association for
  Computational Linguistics.

\bibitem[\protect\citename{Nakashole \bgroup et al.\egroup
  }2012]{Nakashole:2012:PTR:2390948.2391076}
Ndapandula Nakashole, Gerhard Weikum, and Fabian Suchanek.
\newblock 2012.
\newblock Patty: A taxonomy of relational patterns with semantic types.
\newblock In {\em Proceedings of the 2012 Joint Conference on Empirical Methods
  in Natural Language Processing and Computational Natural Language Learning},
  EMNLP-CoNLL '12, pages 1135--1145, Stroudsburg, PA, USA. Association for
  Computational Linguistics.

\bibitem[\protect\citename{Niklaus \bgroup et al.\egroup }2016]{Niklaus2016}
Christina Niklaus, Bernhard Bermeitinger, Siegfried Handschuh, and Andr\'{e}
  Freitas.
\newblock 2016.
\newblock A sentence simplification system for improving relation extraction.
\newblock In {\em Prooceedings of COLING 2016: System Demonstrations, The 26th
  International Conference on Computational Linguistics, Osaka, Japan, December
  11-16, 2016}, pages 170--174.

\bibitem[\protect\citename{Pal and Mausam}2016]{Harinder16}
Harinder Pal and Mausam.
\newblock 2016.
\newblock Demonyms and compound relational nouns in nominal open ie.
\newblock In {\em Proceedings of the 5th Workshop on Automated Knowledge Base
  Construction}, pages 35--39. Association for Computational Linguistics.

\bibitem[\protect\citename{Richardson \bgroup et al.\egroup
  }2013]{richardson2013mctest}
Matthew Richardson, Christopher~JC Burges, and Erin Renshaw.
\newblock 2013.
\newblock Mctest: A challenge dataset for the open-domain machine comprehension
  of text.
\newblock In {\em Proceedings of the 2013 Conference on Empirical Methods in
  Natural Language Processing}, pages 193--203.

\bibitem[\protect\citename{Riloff and
  Jones}1999]{Riloff:1999:LDI:315149.315364}
Ellen Riloff and Rosie Jones.
\newblock 1999.
\newblock Learning dictionaries for information extraction by multi-level
  bootstrapping.
\newblock In {\em Proceedings of the Sixteenth National Conference on
  Artificial Intelligence and the Eleventh Innovative Applications of
  Artificial Intelligence Conference Innovative Applications of Artificial
  Intelligence}, AAAI '99/IAAI '99, pages 474--479, Menlo Park, CA, USA.
  American Association for Artificial Intelligence.

\bibitem[\protect\citename{Saha \bgroup et al.\egroup }2017]{bonie2017}
Swarnadeep Saha, Harinder Pal, and Mausam.
\newblock 2017.
\newblock Bootstrapping for numerical open ie.
\newblock In {\em Proceedings of the 55th Annual Meeting of the Association for
  Computational Linguistics (Volume 2: Short Papers)}, pages 317--323.
  Association for Computational Linguistics.

\bibitem[\protect\citename{Sandhaus}2008]{Sandhaus08}
Evan Sandhaus.
\newblock 2008.
\newblock {The New York Times Annotated Corpus}.
\newblock {\em Linguistic Data Consortium, Philadelphia}, 6(12).

\bibitem[\protect\citename{Schmidek and Barbosa}2014]{Schmid14}
Jordan Schmidek and Denilson Barbosa.
\newblock 2014.
\newblock Improving open relation extraction via sentence re-structuring.
\newblock In {\em Proceedings of the Ninth International Conference on Language
  Resources and Evaluation (LREC-2014)}. European Language Resources
  Association (ELRA).

\bibitem[\protect\citename{Schneider \bgroup et al.\egroup }2017]{Schneider17}
Rudolf Schneider, Tom Oberhauser, Tobias Klatt, Felix~A. Gers, and Alexander
  L{\"o}ser.
\newblock 2017.
\newblock Analysing errors of open information extraction systems.
\newblock In {\em Proceedings of the First Workshop on Building Linguistically
  Generalizable NLP Systems}, pages 11--18. Association for Computational
  Linguistics.

\bibitem[\protect\citename{Stanovsky and Dagan}2016]{Stanovsky2016EMNLP}
Gabriel Stanovsky and Ido Dagan.
\newblock 2016.
\newblock Creating a large benchmark for open information extraction.
\newblock In {\em Proceedings of the 2016 Conference on Empirical Methods in
  Natural Language Processing (EMNLP)}, page (to appear), Austin, Texas,
  November. Association for Computational Linguistics.

\bibitem[\protect\citename{Stanovsky \bgroup et al.\egroup
  }2016]{StanovskyFDG16}
Gabriel Stanovsky, Jessica Ficler, Ido Dagan, and Yoav Goldberg.
\newblock 2016.
\newblock Getting more out of syntax with props.
\newblock {\em CoRR}, abs/1603.01648.

\bibitem[\protect\citename{Surdeanu \bgroup et al.\egroup
  }2003]{surdeanu2003using}
Mihai Surdeanu, Sanda Harabagiu, John Williams, and Paul Aarseth.
\newblock 2003.
\newblock Using predicate-argument structures for information extraction.
\newblock In {\em Proceedings of the 41st Annual Meeting on Association for
  Computational Linguistics-Volume 1}, pages 8--15. Association for
  Computational Linguistics.

\bibitem[\protect\citename{Surdeanu}2013]{surdeanu2013overview}
Mihai Surdeanu.
\newblock 2013.
\newblock Overview of the tac2013 knowledge base population evaluation: English
  slot filling and temporal slot filling.
\newblock In {\em TAC}.

\bibitem[\protect\citename{White \bgroup et al.\egroup
  }2016]{white2016universal}
Aaron~Steven White, Drew Reisinger, Keisuke Sakaguchi, Tim Vieira, Sheng Zhang,
  Rachel Rudinger, Kyle Rawlins, and Benjamin Van~Durme.
\newblock 2016.
\newblock Universal decompositional semantics on universal dependencies.
\newblock In {\em Proceedings of the 2016 Conference on Empirical Methods in
  Natural Language Processing}, pages 1713--1723.

\bibitem[\protect\citename{Wu and Weld}2010]{WuFei10}
Fei Wu and S.~Daniel Weld.
\newblock 2010.
\newblock Open information extraction using wikipedia.
\newblock In {\em Proceedings of the 48th Annual Meeting of the Association for
  Computational Linguistics}, pages 118--127. Association for Computational
  Linguistics.

\bibitem[\protect\citename{Xu \bgroup et al.\egroup }2013]{Xu13}
Ying Xu, Mi-Young Kim, Kevin Quinn, Randy Goebel, and Denilson Barbosa.
\newblock 2013.
\newblock Open information extraction with tree kernels.
\newblock In {\em Proceedings of the 2013 Conference of the North American
  Chapter of the Association for Computational Linguistics: Human Language
  Technologies}, pages 868--877, Atlanta, Georgia, June. Association for
  Computational Linguistics.

\bibitem[\protect\citename{Yahya \bgroup et al.\egroup }2014]{Yahya2014}
Mohamed Yahya, Steven Whang, Rahul Gupta, and Alon Halevy.
\newblock 2014.
\newblock Renoun: Fact extraction for nominal attributes.
\newblock In {\em Proceedings of the 2014 Conference on Empirical Methods in
  Natural Language Processing (EMNLP)}, pages 325--335. Association for
  Computational Linguistics.

\bibitem[\protect\citename{Zhang \bgroup et al.\egroup
  }2017]{zhang-EtAl:2017:IWCS}
Sheng Zhang, Rachel Rudinger, and Ben Van~Durme.
\newblock 2017.
\newblock An evaluation of predpatt and open ie via stage 1 semantic role
  labeling.
\newblock In {\em Proceedings of the 12th International Conference on
  Computational Semantics (IWCS)}, Montpellier, France, September.

\end{thebibliography}

\end{document}